\documentclass[sigconf]{acmart}

\usepackage{enumerate}
\usepackage{enumitem}
\usepackage{multirow}
\usepackage{bbding}
\usepackage{pifont}
\usepackage{wasysym}
\usepackage{amssymb}
\usepackage{bm}
\usepackage{graphicx}
\usepackage{amsmath}
\usepackage{float}
\usepackage{color}
\usepackage{subfigure}
\usepackage[normalem]{ulem}
\useunder{\uline}{\ul}{}
\acmSubmissionID{267}

\copyrightyear{2023}
\acmYear{2023}
\setcopyright{acmcopyright}\acmConference[WSDM '23]{Proceedings of the Sixteenth ACM International Conference on Web Search and Data Mining}{February 27-March 3, 2023}{Singapore, Singapore}
\acmBooktitle{Proceedings of the Sixteenth ACM International Conference on Web Search and Data Mining (WSDM '23), February 27-March 3, 2023, Singapore, Singapore}
\acmPrice{15.00}
\acmDOI{10.1145/3539597.3570405}
\acmISBN{978-1-4503-9407-9/23/02}

\begin{document}

\title{Multi-queue Momentum Contrast for Microvideo-Product Retrieval}
\titlenote{This research was partially supported by NSFC (62076121)}

\author{Yali Du}
\affiliation{%
  \institution{Nanjing University}
  \country{}}
\email{duyali2000@gmail.com}

\author{Yinwei Wei}
\authornote{Yinwei Wei is the corresponding author.}
\affiliation{%
  \institution{National University of Singapore}
  \country{}}
\email{weiyinwei@hotmail.com}

\author{Wei Ji}
\affiliation{%
  \institution{National University of Singapore}
  \country{}}
\email{jiwei@nus.edu.sg}

\author{Fan Liu}
\affiliation{%
  \institution{National University of Singapore}
  \country{}}
\email{liufancs@gmail.com}
  
\author{Xin Luo}
\affiliation{%
  \institution{Shandong University}
  \country{}}
\email{luoxin.lxin@gmail.com}

\author{Liqiang Nie}
\affiliation{%
  \institution{Harbin Institute of Technology (Shenzhen)}
  \country{}}
\email{nieliqiang@gmail.com}
  
\renewcommand{\shortauthors}{Yali Du, Yinwei Wei, Wei Ji, Fan Liu, Xin Luo, \& Liqiang Nie}
  
\begin{abstract}
The booming development and huge market of micro-videos bring new e-commerce channels for merchants. Currently, more micro-video publishers prefer to embed relevant ads into their micro-videos, which not only provides them with business income but helps the audiences to discover their interesting products. 
However, due to the micro-video recording by unprofessional equipment, involving various topics and including multiple modalities, it is challenging to locate the products related to micro-videos efficiently, appropriately, and accurately.
We formulate the microvideo-product retrieval task, which is the first attempt to explore the retrieval between the multi-modal and multi-modal instances.

A novel approach named Multi-Queue Momentum Contrast (MQMC) network is proposed for bidirectional retrieval, consisting of the uni-modal feature and multi-modal instance representation learning. Moreover, a discriminative selection strategy with a multi-queue is used to distinguish the importance of different negatives based on their categories. 
We collect two large-scale microvideo-product datasets (MVS and MVS-large) for evaluation and manually construct the hierarchical category ontology, which covers sundry products in daily life. Extensive experiments show that MQMC outperforms the state-of-the-art baselines.
Our replication package (including code, dataset, etc.) is publicly available at \url{https://github.com/duyali2000/MQMC}.

\end{abstract}



\maketitle

\section{Introduction}
Micro-video, as a new form of social media, has become an important component in our daily life. Compared with long videos, micro-videos have the instincts of short duration and easy-to-share, making them popular in online sharing platforms. Taking Tiktok as an example, its Monthly Active Users had reached one billion until September 2021\footnote{https://new.qq.com/omn/20210928/20210928A01WW400.html}. Such a huge market undoubtedly brings new E-commerce chance for merchants. Currently, some publishers start to embed relevant ads into their micro-videos, which not only provides them external income but facilitates the audiences to discover their interested products\cite{micro1,micro2,micro3,micro4,micro6,semi,liu2017towards,liu2018online,wei2019mmgcn,micro7,xin2022user}. 

However, the irrelevant products may harm the audiences' experience and make them disappointed with the micro-video itself. Hence, how to locate the products related to micro-videos challenges micro-video publishers. 
To remedy this problem, it is of great importance to understand the semantic information of the micro-video and product, so as to measure their affinities. The prior studies hence frame this problem as the video-shop retrieval task. 
For instance, Zhao \textit{et. al.,}~\cite{dress} proposed a DPRnet model to discover the keyframe from videos and measure visual similarities with other candidate items. 
Considering the temporal relation hidden in the video, Cheng \textit{et. al.,}~\cite{video2shop} applied the LSTM model to represent the video and developed an AsymNet model to discover the matched items according to their distances to the video. More recently, Godi \textit{et. al.,}~\cite{movingfasion} constructed a new dataset composed of social videos and their corresponding products' images, proposed SEAM Match-RCNN to extract features from video, and located the visually similar clothes. 

Despite their remarkable performance, it is hard to directly adopt these methods to seek the relevant products for micro-videos, due to the following problem:
\begin{itemize}[leftmargin=*]
\item  The micro-video tends to be shot by amateurs with some non-professional equipment, like a mobile phone and pad. Thereby, the micro-video inevitably contains some environmental noise, resulting in sub-optimal representations of micro-videos. 
\item Unlike the videos shot by merchants, the micro-video is not designed for the specific product. Hence, the micro-video may involve various topics, which is more challenging than traditional video-shop retrieval. 
\item The micro-video and product embrace the signal from the multiple modalities, including the visual and textual cues. Therefore, not only cross-modal but intra-modal relations should be considered in microvideo-product matching. 
\end{itemize}

To resolve this problem, we resort to address two technical challenges: 
(1) \textit{how to distill the informative signal relevant to the product from the content information of micro-video}, and (2) \textit{how to represent the multi-modal micro-video and product to model their similarities}.

Therefore, we develop a new microvideo-product retrieval model, termed Multi-Queue Momentum Contrast (MQMC) method, which consists of the uni-model feature representation learning and multi-modal instance representation learning. 
To optimize the uni-modal feature encoders, we introduce cross-modal contrastive loss and intra-modal contrastive loss. 
The former models the consistency between the visual and textual modalities while the latter utilizes the supervision signal from the product information to help the encoders to distinguish the informative signal from the irrelevant content of the micro-video.
In multi-modal instance representation learning, we resort to the
momentum-based contrastive loss\cite{he2020momentum} to model the instance-level similarity. Considering different negative micro-videos (products) play various importance to the anchor products (micro-videos), we take a negative selection strategy with multi-queue to distinguish the importance of different negatives by measuring the distance of categories of anchor and negatives.

To evaluate our proposed model, we collect the microvideo-product pairs from the popular micro-video sharing platforms and achieve two datasets: MVS and MVS-large,  which contain $13,165$ and $126,206$ microvideo-product pairs respectively. In addition, we manually construct the hierarchical category ontology including $6$ upper ontologies, $30$ middle ontologies, and $316$ lower ontologies. By conducting extensive experiments on these two datasets, MQMC significantly outperforms the state-of-the-art methods, which demonstrates the effectiveness of our proposed model.

In a nutshell, our contributions could be summarized as follows:
\begin{itemize}[leftmargin=*]
\item By investigating the prior studies on cross-modal information retrieval, we formulate a new microvideo-product retrieval task. To the best of our knowledge, this is the first attempt to explore the retrieval between the multi-modal and multi-modal instances.
\item We propose a novel Multi-Queue Momentum Contrast (MQMC) network consisting of the uni-modal feature and multi-modal instance representation learning, so as to locate the relevant micro-video (product) for the inputted product (micro-video). 
\item We design a new multi-queue contrastive training strategy, which maintains multiple queues of negative samples and considers the importance of different negatives based on their categories in contrastive loss computation. 
\item To evaluate our proposed model, we construct two large-scale datasets \textit{i.e.,} MVS and MVS-large and build the hierarchical category ontology of the products. By conducting extensive experiments on the datasets, we demonstrate that our proposed model outperforms the state-of-the-art baselines by a margin. 
\end{itemize}

\section{Related Work}
\subsection{Cross-modal information retrieval}
With the explosion of multi-modal data, cross-modal retrieval has attracted wide attention, mainly focusing on image-text, video-text, video-image, etc\cite{clip,hit,ALBEF,avlnet,howto100m,li2009exploiting,cross-modal1,ye2016college,yang2021corporate,liu2017towards,liu2018online,SunWJCSN22,SunWSFN22,wei2019mmgcn,liu2022privacy,hu2021coarse}.
CLIP\cite{clip} used a sufficiently large dataset for pre-training and natural language as a supervisory signal to learn visual representation. 
ALBEF\cite{ALBEF} introduced a contrastive loss to align the image and text representations before fusing them through cross-modal attention. 
Hit\cite{hit} combined hierarchical transformer and momentum contrast method for video-text retrieval. MMT\cite{mmt} learned an effective representation from different modalities inherent in video over multiple self-attention layers with several video feature extractors. Most of all, the above methods are based on the tasks of single-modal to single-modal, or single-modal to multi-modal, which do not apply to the task of multi-modal to multi-modal retrieval in our paper.

The recent influx of instructional multi-modal datasets such as Inria Instructional Videos\cite{Inria}, CrossTask\cite{CrossTask}, YouCook2\cite{YouCook2}, and HowTo100M\cite{howto100m} has inspired a variety of methods for video-text retrieval, but those are not suitable for the task of microvideo-product retrieval in this paper. AsymNet\cite{video2shop}, DPRNet\cite{dress}, and Fashion Focus\cite{zhang2021fashion} built video-to-shop datasets from advertisements in online shopping platforms, but the datasets are not publicly released. 
Although MovingFashion\cite{godi2022movingfashion} and Watch and Buy\cite{rao2021watch} are
publicly available datasets of "video-to-shop", they have the disadvantage of a single domain (only clothing).

\begin{figure*}
\vspace{-0.2cm}
    \setlength{\abovecaptionskip}{0.2cm}
    \setlength{\belowcaptionskip}{0.cm}
\centering
\includegraphics[width=\textwidth]{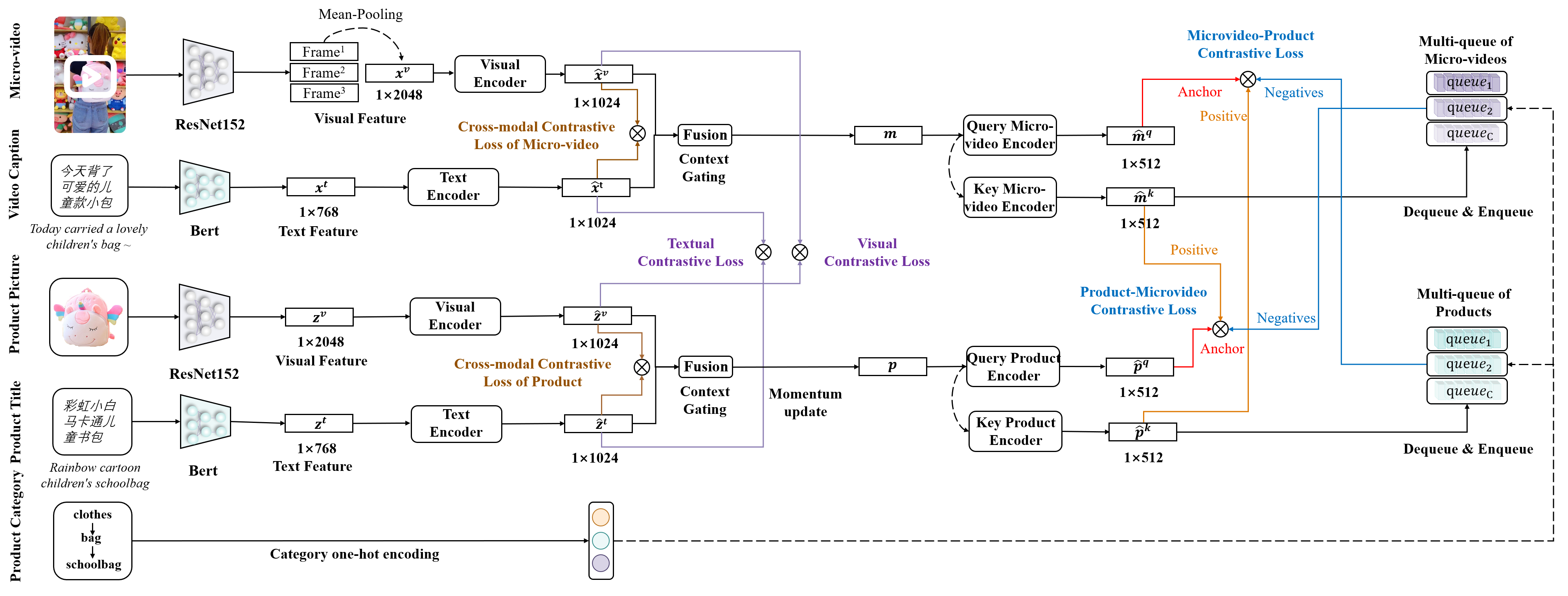}
\caption{An overview of MQMC for microvideo-product retrieval.}
\label{mqmc}
\end{figure*}

\subsection{Contrastive Learning}

Contrastive learning is widely studied in self-supervised and unsupervised learning and has made many remarkable achievements\cite{he2020momentum,simclr,contrastive1,contrastive2,cmc,contrastive3,infonce,contrastive4,infonce,contrastive6,contrastive8}. The contrastive learning model is built on the principle that positives are closer to each other in the projection space, while negatives are farther apart. The main challenges are how to choose positives and negatives, how to construct a representation learning model that can follow the above principle, and how to prevent model collapse.

According to the way of choosing negative samples, current contrastive learning methods can be divided into in-batch and out-batch. 
The end-to-end mechanism uses the anchor’s augmented views as positives and considers other samples in the current batch as negatives. SimCLR\cite{simclr} achieved success in unsupervised visual representation learning, which benefits from large batch size, stronger data augmentation, and the learnable nonlinear projection head.
The memory bank mechanism constructs a memory bank to memorize broader negative samples, which has the drawback that the samples in the memory bank are from very different encoders all over the past epoch and they are less consistent. Faced with this problem, MoCo\cite{he2020momentum} used a momentum-updated key encoder to maintain the consistency of negative representations in the memory bank. 

However, increasing the memory size or batch size does not always improve the performance rapidly, because more negatives do not necessarily mean that more difficult negatives are brought. There are many recent improvements in contrastive learning, including loss function, sampling strategy, and data augmentation, but few relative works on negative samples. The existing works include making use of labels and generating difficult negative samples through Mixup\cite{hard,hardneg}.

Our method in multi-modal microvideo-product retrieval benefits from the large-scale negatives using a memory bank and the multi-layer category ontology to distinguish the hard negatives.

\section{Method}
In this section, we first formulate microvideo-product retrieval task, and then detail our proposed model, as shown in Figure~\ref{mqmc}, consisting of the uni-modal feature representation learning and multi-modal instance representation learning.
\subsection{Problem Definition}
Given a set $\mathcal{V}=\{v_1, v_2, \dots, v_M\}$ of $M$ micro-video, the microvideo-product retrieval task aims to discover the most similar product from a set $\mathcal{P}=\{p_1, p_2, \dots, p_N\}$ of $N$ products. More specifically, for the $i$-th micro-video, we use a pre-trained ResNet model~\cite{resnet152} to extract the visual features from the keyframes. With these features, we perform a mean-pooling operation to obtain the visual feature vector of micro-video $\mathbf{x}^v_i\in\mathbb{R}^{d_v}$, where $d_v$ is the dimension of the visual features\footnote{ 
Without any particular clarification, all the vectors are in column forms.}. Beyond the visual signal, we consider the caption of $i$-th micro-video and capture its textual features with the trained BERT model~\cite{bert}, denoted as $\mathbf{x}^t_i\in\mathbb{R}^{d_t}$. 
Analogously, we obtain the visual and textual features from the images and descriptions of products with the same extractors. Taking $j$-th product as an example, we denote its visual and textual feature vector as $\mathbf{z}^v_j$ and $\mathbf{z}^t_j$, respectively. 
Formally, with the obtained features of micro-videos and products, we aim to learn a function to score the similarities of microvideo-product pairs:
\begin{equation}
\vspace{-0.1cm}
    s_{i,j} = f(\mathbf{x}^v_i, \mathbf{x}^t_i; \mathbf{z}^v_j, \mathbf{z}^t_j), 
\end{equation}
where $f(\cdot)$ is the similarity scoring function and $s_{i,j}$ represents the similarity between the $i$-th micro-video and $j$-th product. 


\subsection{Framework}
\subsubsection{\textbf{Uni-modal Feature Representation Learning}}
As aforementioned, the visual and textual content of the micro-video contains noise signals, like the surroundings and emotional expression, which are irrelevant to the target product search. Hence, we introduce uni-modal feature encoders for two modalities and design the cross- and intra-modal loss functions to distill the informative signal. 
\par\noindent\textbf{Uni-modal Feature Encoder}. 
With the obtained visual and textual features, we separately adopt the two-layer neural networks equipped with a nonlinear activation function on two modalities:
\begin{equation}
\vspace{-0.1cm}
    \begin{cases}
   \hat{\mathbf{x}}^{v}_i =  \bm{W}^v_2\phi(\bm{W}_1^v\mathbf{x}_i^v),\\
    \hat{\mathbf{x}}^{t}_i = \bm{W}^t_2\phi(\bm{W}_2^t\mathbf{x}_i^t),
    \end{cases}
\end{equation}
where $\bm{W}^v_{(\cdot)}$ and $\bm{W}^t_{(\cdot)}$ are the trainable parameters in visual and textual modalities, respectively. And, $\phi(\cdot)$ is the $leaky\_relu$ function~\cite{leakyrelu} in our experiments. $\hat{\mathbf{x}}^{v}_i\in\mathbb{R}^{d}$ and $\hat{\mathbf{x}}^{t}_i\in\mathbb{R}^{d}$ are the refined vectors in visual and textual modalities, respectively. Wherein, $d$ is the dimension of the mapped spaces. Note that we omit the bias term for briefness.

Moreover, we also implement a two-layer network to map the visual (textual) features of the product into the same space of the refined visual (textual) vector, \textit{i.e.,} $\hat{\mathbf{z}}^{v}_i\in\mathbb{R}^{d}$ and $\hat{\mathbf{z}}^{t}_i\in\mathbb{R}^{d}$. These mapping operations are conducted for the following two contrastive loss functions. 

\noindent\textbf{Cross-modal Contrastive Loss}.
To optimize the feature encoders, we introduce a cross-modal contrastive loss to explicitly model the consistency between the visual and textual. 
Specifically, it is implemented by identifying the positive pair of cross-modal vectors, like <$\hat{\mathbf{x}}^{v}_i, \hat{\mathbf{x}}^{t}_i$>, learned from the same instance (\textit{i.e.,} micro-video or product) from multiple negative pairs of the vectors from different instances. To be more specific, we treat $\hat{\mathbf{x}}^v_i$ as the anchor and $\hat{\mathbf{x}}^t_i$ as the positive vector. And, we randomly sample multiple textual vectors of other micro-videos as the negative vectors. Similarity, we establish the cross-modal negative pairs for the product, formally,
\begin{equation}
\vspace{-0.1cm}
       \{\hat{\mathbf{x}}^{t}_{i,1}, \hat{\mathbf{x}}^{t}_{i,2}, \dots, \hat{\mathbf{x}}^{t}_{i,K}\}; \ \ 
    \{\hat{\mathbf{z}}^{t}_{i,1}, \hat{\mathbf{z}}^{t}_{i,2}, \dots, \hat{\mathbf{z}}^{t}_{i,K}\},  
\end{equation}
where $\hat{\mathbf{x}}^{t}_{i,\cdot}$ and $\hat{\mathbf{z}}^{t}_{j,\cdot}$ are the negative vectors of $i$-th micro-video and $j$-th product. And, $K$ is the pre-defined number of negative vectors.

With the anchor, positive, and negative vectors, we opt for the InfoNCE\cite{infonce} loss, formally,
\begin{equation}
\vspace{-0.1cm}
    \begin{split}
    \mathcal{L}_1 = -\ln\frac{\exp(\frac{(\hat{\mathbf{x}}_i^v)^\top\hat{\mathbf{x}}_i^t}{||\hat{\mathbf{x}}_i^v||\cdot||\hat{\mathbf{x}}_i^t||}\cdot\frac{1}{\tau})}{\exp(\frac{(\hat{\mathbf{x}}_i^v)^\top\hat{\mathbf{x}}_i^t}{||\hat{\mathbf{x}}_i^v||\cdot||\hat{\mathbf{x}}_i^t||}\cdot\frac{1}{\tau})+\sum\limits_{k=1}^{K}\exp(\frac{(\hat{\mathbf{x}}_i^v)^\top\hat{\mathbf{x}}_{i,k}^t}{||\hat{\mathbf{x}}_i^v||\cdot||\hat{\mathbf{x}}_{i,k}^t||}\cdot\frac{1}{\tau})} \\-\ln\frac{\exp(\frac{(\hat{\mathbf{z}}_i^v)^\top\hat{\mathbf{z}}_i^t}{||\hat{\mathbf{z}}_i^v||\cdot||\hat{\mathbf{z}}_i^t||}\cdot\frac{1}{\tau})}{\exp(\frac{(\hat{\mathbf{z}}_i^v)^\top\hat{\mathbf{z}}_i^t}{||\hat{\mathbf{z}}_i^v||\cdot||\hat{\mathbf{z}}_i^t||}\cdot\frac{1}{\tau})+\sum\limits_{k=1}^{K}\exp(\frac{(\hat{\mathbf{z}}_i^v)^\top\hat{\mathbf{z}}_{i,k}^t}{||\hat{\mathbf{z}}_i^v||\cdot||\hat{\mathbf{z}}_{i,k}^t||}\cdot\frac{1}{\tau})},
    \end{split}
\end{equation}
where $\tau$ is a temperature parameter.

\noindent\textbf{Intra-modal Contrastive Loss}. 
Beyond the consistency between different modalities, we further leverage the supervision signal from the product information to optimize the feature encoders of the micro-video. It can help the encoders to distinguish the informative signal from the irrelevant content of the micro-video. Following similar operations, we construct the contrastive pairs of the micro-video and product in each modality. In particular, we group the $i$-th micro-video and corresponding product as the positive pair, \textit{i.e.,} $<\hat{\mathbf{x}}^{v}_i, \hat{\mathbf{z}}^{v}_i>$ and $<\hat{\mathbf{x}}^{t}_i, \hat{\mathbf{z}}^{t}_i>$. And then, we randomly sample multiple products as the negative samples:
\begin{equation}
\vspace{-0.1cm}
    \{\hat{\mathbf{z}}^{v}_{i,1}, \hat{\mathbf{z}}^{v}_{i,2}, \dots, \hat{\mathbf{z}}^{v}_{i,K}\},\ \  
    \{\hat{\mathbf{z}}^{t}_{i,1}, \hat{\mathbf{z}}^{t}_{i,2}, \dots, \hat{\mathbf{z}}^{t}_{i,K}\}. 
\end{equation}

Accordingly, we conduct the contrastive loss, formally, 
\begin{equation}
\vspace{-0.1cm}
    \begin{split}
    \mathcal{L}_2 = -\ln\frac{\exp(\frac{(\hat{\mathbf{x}}_i^v)^\top\hat{\mathbf{z}}_i^v}{||\hat{\mathbf{x}}_i^v||\cdot||\hat{\mathbf{z}}_i^v||}\cdot\frac{1}{\tau})}{\exp(\frac{(\hat{\mathbf{x}}_i^v)^\top\hat{\mathbf{z}}_i^v}{||\hat{\mathbf{x}}_i^v||\cdot||\hat{\mathbf{z}}_i^v||}\cdot\frac{1}{\tau})+\sum\limits_{k=1}^{K}\exp(\frac{(\hat{\mathbf{x}}_i^v)^\top\hat{\mathbf{z}}_{i,k}^v}{||\hat{\mathbf{x}}_i^v||\cdot||\hat{\mathbf{z}}_{i,k}^v||}\cdot\frac{1}{\tau})} \\-\ln\frac{\exp(\frac{(\hat{\mathbf{x}}_i^t)^\top\hat{\mathbf{z}}_i^t}{||\hat{\mathbf{x}}_i^t||\cdot||\hat{\mathbf{z}}_i^t||}\cdot\frac{1}{\tau})}{\exp(\frac{(\hat{\mathbf{x}}_i^t)^\top\hat{\mathbf{z}}_i^t}{||\hat{\mathbf{x}}_i^t||\cdot||\hat{\mathbf{z}}_i^t||}\cdot\frac{1}{\tau})+\sum\limits_{k=1}^{K}\exp(\frac{(\hat{\mathbf{x}}_i^t)^\top\hat{\mathbf{z}}_{i,k}^t}{||\hat{\mathbf{x}}_i^t||\cdot||\hat{\mathbf{z}}_{i,k}^t||}\cdot\frac{1}{\tau})}.
    \end{split}
\end{equation}

\subsubsection{\textbf{Multi-modal Instance Representation Learning}}
After obtaining the refined features of the micro-video and product, we aim to fuse the multi-modal feature to capture multi-modal instance representations. In this part, we elaborate on the fusion model and cross-instance contrastive loss to optimize it. 
\par\noindent\textbf{Multi-modal Fusion}. We apply the context gating mechanism~\cite{gating} to fuse the visual and textual features of instances, including micro-videos and products. It is defined as 
\begin{equation}
\vspace{-0.1cm}
\begin{cases}
    \bm{m}_i = (\bm{W}_2^m\hat{\mathbf{x}}^v_{i} + \bm{W}_1^m\hat{\mathbf{x}}^t_{i}) \circ 
    \sigma (\bm{W}_3^m(\bm{W}_2^m\hat{\mathbf{x}}^v_{i} + \bm{W}_1^m\hat{\mathbf{x}}^t_{i})),\\

    \bm{p}_j = (\bm{W}_2^p\hat{\mathbf{z}}^v_{j} + \bm{W}_1^p\hat{\mathbf{z}}^t_{j}) \circ 
    \sigma (\bm{W}_3^p(\bm{W}_2^p\hat{\mathbf{z}}^v_{j} + \bm{W}_1^p\hat{\mathbf{z}}^t_{j})),
\end{cases}
\end{equation}
where $\bm{m}_i\in\mathbb{R}^{d'}$ and $\bm{p}_j\in\mathbb{R}^{d'}$ are the multi-model feature vectors of $i$-th micro-video and $j$-th product, respectively. $d'$ is the dimension of the multi-model feature vector. Moreover, $\bm{W}_{\cdot}^m$ and $\bm{W}_{\cdot}^p$ are the trainable weight matrices in the micro-video and product fusion models. 
$\circ$ denotes element-wise multiplication and $\sigma$ is an element-wise sigmoid activation. 

\begin{figure}
\vspace{-0.2cm}
    \setlength{\abovecaptionskip}{0.2cm}
    \setlength{\belowcaptionskip}{0.cm}
\centering
\includegraphics[width=1.5in]{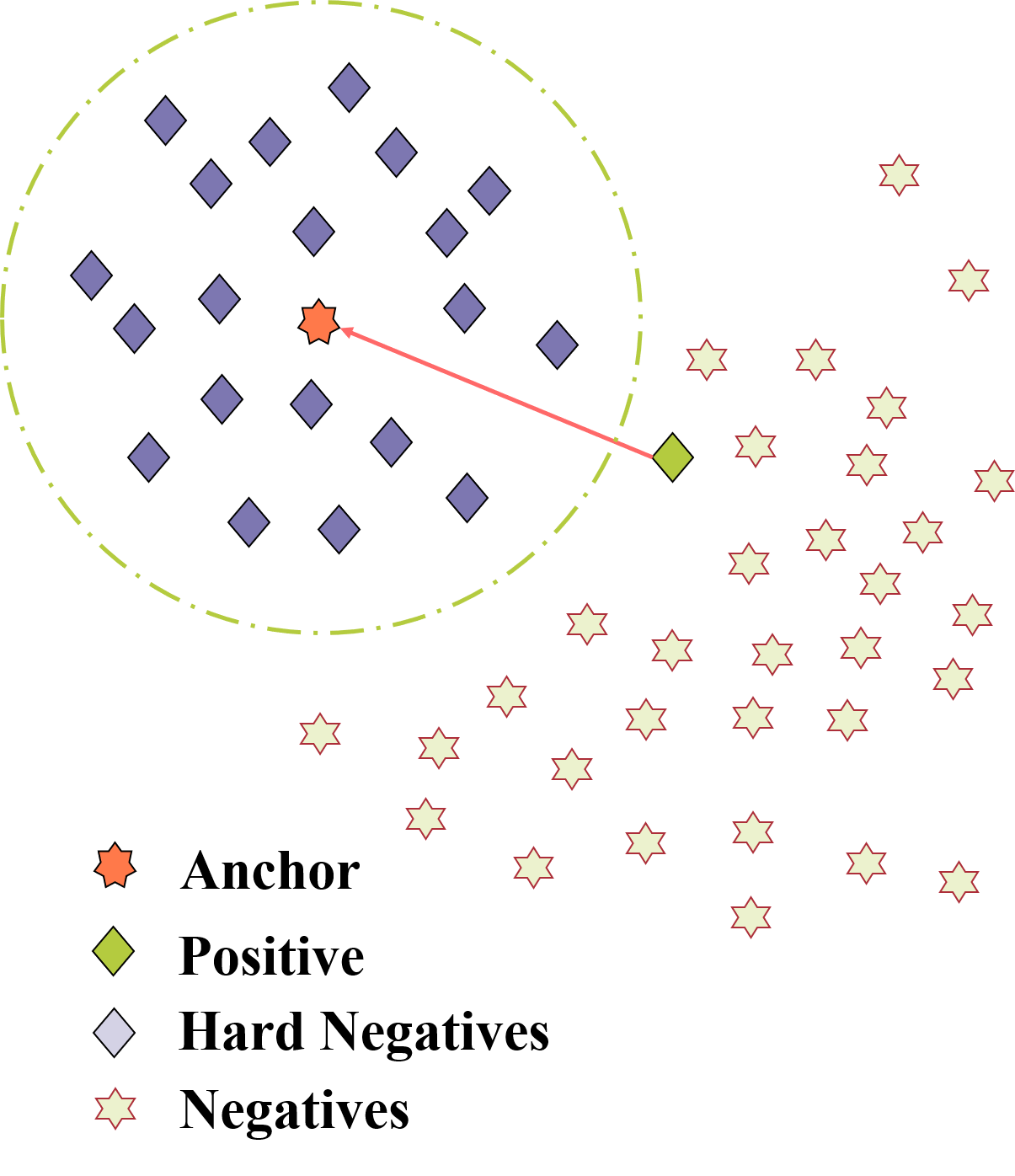}
\caption{Importance of negatives in different categories.}
\label{importance}
\end{figure}

\noindent\textbf{Cross-instance Contrastive Loss}. 
Facing a vast number of candidate micro-videos and products in practice, we resort to the momentum-based contrastive loss~\cite{he2020momentum} to model the instance-level similarity. 
Typically, a queue is maintained to store the encoded representations of samples. With the momentum update mechanism for parameters and the memory-bank update mechanism of enqueuing the new mini-batch samples and dequeuing the oldest ones, we can obtain large and consistent instances to conduct contrastive learning stably and smoothly. 

Towards this end, we build the query and key encoders for the micro-video and product and feed the representations of instances into them, formally 
\begin{equation}
\vspace{-0.1cm}
    \begin{cases}
    \hat{\bm{m}}^k_i = h(\bm{m}_i; {\theta^k_m}),\\
    \hat{\bm{m}}^q_i = h(\bm{m}_i; {\theta^q_m}),\\
    \hat{\bm{p}}^k_j = h(\bm{p}_j; {\theta^k_p}),\\
    \hat{\bm{p}}^q_j = h(\bm{p}_j; {\theta^q_p}).\\
    \end{cases}
\end{equation}
Wherein, $k$ and $q$ are used to indicate the key and query encoders, respectively. 
$\hat{\bm{m}}^{(\cdot)}_i\in\mathbb{R}^{d'}$ and $\hat{\bm{p}}^{(\cdot)}_j\in\mathbb{R}^{d'}$ separately denote the encoded vectors of micro-video and product. $h(\cdot)$ denotes the encoder, in which $\theta^{(\cdot)}_m$ and $\theta^{(\cdot)}_p$ are the parameters to be trained. 

Based on the momentum contrastive learning mechanism, the parameters $\theta^q_{m}$ and $\theta^q_{p}$ are updated by back-propagation, while $\theta^k_{m}$ and $\theta^k_{p}$ are momentum updated as follows:
\begin{equation}
\vspace{-0.15cm}
\begin{cases}
\theta^k_{m} \gets m\theta^k_{m} + (1 - m)\theta^q_{m},\\
\theta^k_{p} \gets m\theta^k_{p} + (1 - m)\theta^q_{p},
\end{cases}
\end{equation}
where $m \in$ $[0, 1)$ is a pre-defined momentum coefficient. 

However, this method seldom considers the difference between the samples in the queue and treats them as the negative samples equally in the contrastive loss computation. In fact, different negative micro-videos (products) play various importance to the anchor products (micro-videos). 
To illustrate this, we take one micro-video and its corresponding product as an anchor and positive instances, and randomly sample multiple products as negative ones. We scatter these instances with t-SNE algorithm~\cite{van2008visualizing}, as illustrated in Figure~\ref{importance}. Observing the distances between the anchor and other instances, we find that a portion of negative samples, namely the `hard' negative sample, are closer to the anchor than the positive one. They make more contributions to optimize the representation learning of instances, whereas the other negative samples hardly help the optimization. 
As the prior study~\cite{hardneg} mentioned, the effective negative in contrastive loss should satisfy the two principles: 
\begin{itemize}[leftmargin=*]
\item 
\textit{Principle 1. The labels of true negatives should differ from that of the anchor x.}
\item 
\textit{Principle 2. The most useful negative samples are ones that the embedding currently believes to be similar to the anchor.}
\end{itemize}

Therefore, we design a multi-queue momentum contrast training strategy by which we organize the mini-batch instances according to their categories (\textit{e.g.,} bag, toy, and fruit) and enqueue them into several category-aware queues. When the anchor and positive instances emerge in the training phase, we dequeue the oldest samples from the queues as the negative samples to perform the momentum contrastive learning. Formally, we build the multi-queue structure as,
\begin{equation}
\vspace{-0.15cm}
\begin{cases}
    queue_1=[q^1_1, q^1_2, \dots, q^1_T],\\
    queue_2=[q^2_1, q^2_2, \dots, q^2_T],\\
    \cdots,\\
    queue_C=[q^C_1, q^C_2, \dots, q^C_T],\\
\end{cases}
\end{equation}
where $queue_t$ denotes $t$-th queue corresponding to $t$-th category. $C$ and $T$ are the numbers of queues and the length of each queue, respectively. 

When a mini-batch of microvideo-product pairs is fed, we collect the negative samples from the multiple queues according to the categories of the positive instances. For instance, given four products that belong to $1$-th, $1$-th, $2$-th, and $3$-th categories, we dequeue two negative samples from $queue_1$, one sample from $queue_2$, and one sample from $queue_3$. 
Reorganizing these dequeued instances as a negative batch, we can guarantee the `hard' negative sample of each anchor in the contrastive training phase. 

Moreover, considering the various categories of instances in the negative batch, we score the importance of anchor and negative sample pairs to formulate a multi-queue contrastive loss. 
For this purpose, we perform a mean-pooling operation on all instances of the same category as the representation of the category. Then, we can score the importance of instance pairs by measuring the distance of representation of their categories, formally, 
\begin{equation}
\vspace{-0.15cm}
    e_{i, j} = 1 - \sum_{l=1}^{L}\zeta\cdot\exp(norm(d(\bm{c}_i^1, \bm{c}_j^1))) - \zeta\cdot\exp(norm(d(\bm{c}_i^2, \bm{c}_j^2))),
\end{equation}
where $e_{i, j}$ denotes the importance score for $i$-th and $j$-th instances. $norm(\cdot)$ and $d(\cdot)$ represent the normalization and Euclidean distance functions, respectively. In addition, $\zeta$ is the hyper-parameter controlling the degree of the score. It is worth noting that since we construct a hierarchical ontology of categories, each instance can be classified into two categories in the fine-grained and coarse-grained levels, \textit{e.g.,} $\mathbf{c}_i^1, \mathbf{c}_j^1\in\mathbb{R}^{d'}$ and $\mathbf{c}_i^2, \mathbf{c}_j^2\in\mathbb{R}^{d'}$. Hence, we compute the similarities of category pairs at two levels.

\begin{equation}
\vspace{-0.1cm}
    \begin{split}
    \mathcal{L}_3 = -\ln\frac{\exp(\frac{(\hat{\mathbf{m}}_i^k)^\top\hat{\mathbf{p}}_i^q}{||\hat{\mathbf{m}}_i^k||\cdot||\hat{\mathbf{p}}_i^q||}\cdot\frac{1}{\tau})}{\exp(\frac{(\hat{\mathbf{m}}_i^k)^\top\hat{\mathbf{p}}_i^q}{||\hat{\mathbf{m}}_i^k||\cdot||\hat{\mathbf{p}}_i^q||}\cdot\frac{1}{\tau})+\sum\limits_{k=1}^{K}e_{i,i_k}\exp(\frac{(\hat{\mathbf{m}}_i^k)^\top\hat{\mathbf{p}}_{i,k}^q}{||\hat{\mathbf{m}}_i^k||\cdot||\hat{\mathbf{p}}_{i,k}^q||}\cdot\frac{1}{\tau})} \\-\ln\frac{\exp(\frac{(\hat{\mathbf{p}}_i^k)^\top\hat{\mathbf{m}}_i^q}{||\hat{\mathbf{p}}_i^k||\cdot||\hat{\mathbf{m}}_i^q||}\cdot\frac{1}{\tau})}{\exp(\frac{(\hat{\mathbf{p}}_i^k)^\top\hat{\mathbf{m}}_i^q}{||\hat{\mathbf{p}}_i^k||\cdot||\hat{\mathbf{m}}_i^q||}\cdot\frac{1}{\tau})+\sum\limits_{k=1}^{K}e_{i,i_k}\exp(\frac{(\hat{\mathbf{p}}_i^k)^\top\hat{\mathbf{m}}_{i,k}^q}{||\hat{\mathbf{p}}_i^k||\cdot||\hat{\mathbf{m}}_{i,k}^q||}\cdot\frac{1}{\tau})}.
    \end{split}
\end{equation}

Thus, the overall objective function is $\mathcal{L}$:
\begin{equation}
\vspace{-0.1cm}
     \mathcal{L} = \alpha \mathcal{L}_1 + \beta \mathcal{L}_2 + \delta \mathcal{L}_3,
\end{equation}
where $\alpha$, $\beta$, and $\delta$ are three hyper-parameters to balance the cross-modal, cross-instance, and cross-category contrasts, restricted in the range of zero to one. We test different compositions of $\alpha$, $\beta$ and $\delta$ in our experiments.

\begin{figure}
\centering
\includegraphics[width=2.5in]{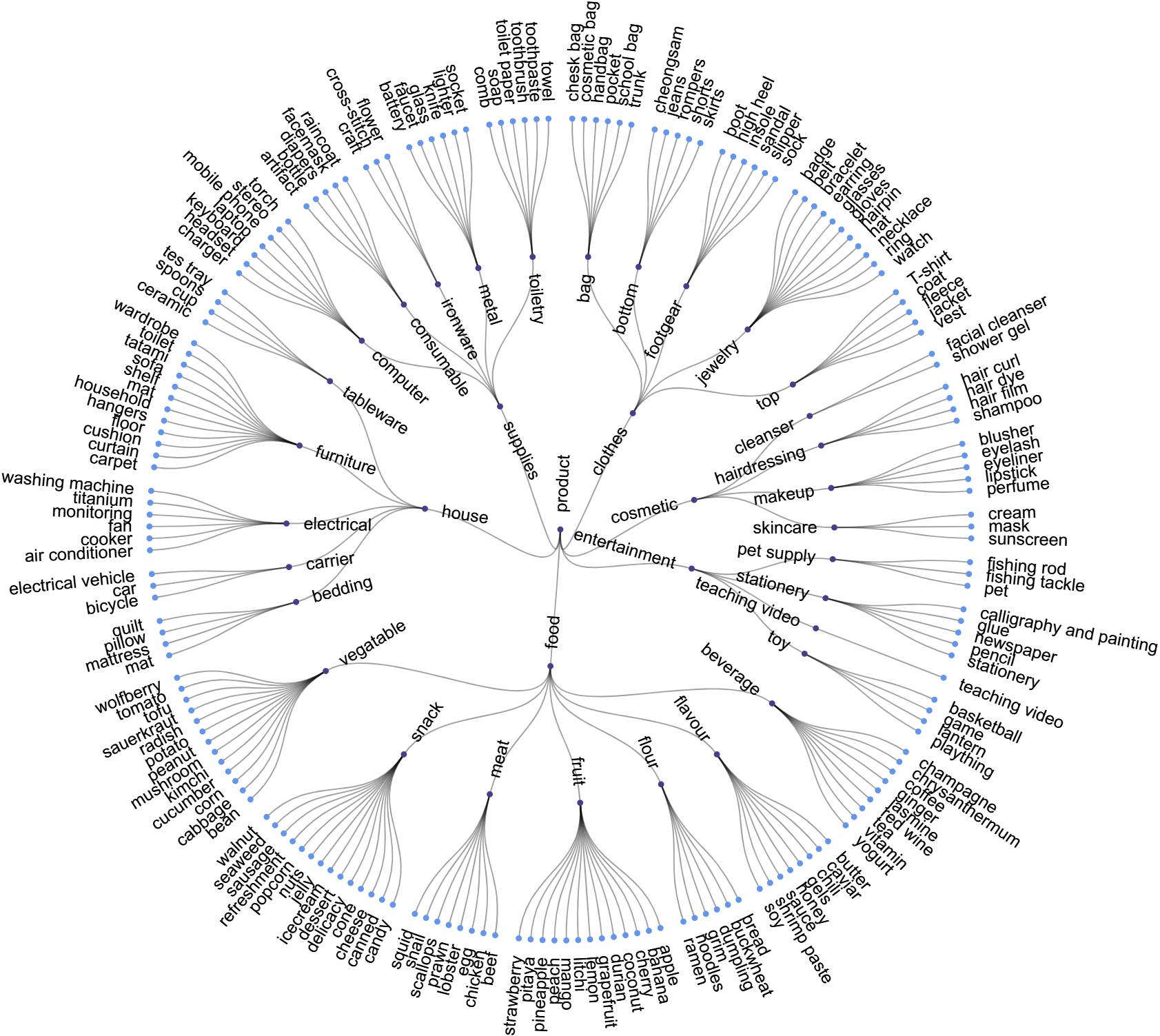}
\caption{The category ontology.}
\label{fig2}
\end{figure}


\begin{table*}
\vspace{-0.2cm}
    \setlength{\abovecaptionskip}{0.2cm}
    \setlength{\belowcaptionskip}{0.cm}
\centering
\caption{The experimental results of start-of-the-art on MVS and MVS-large.}
\label{SOTA} 
\begin{tabular}{c|c|ccccc|ccccc}
\toprule
\multirow{2}{*}{\textbf{Datasets}}
& \multirow{2}{*}{\textbf{Methods}} 
& \multicolumn{5}{c|}{\textbf{Microvideo-Product Retrieval}} 
& \multicolumn{5}{c}{\textbf{Product-Microvideo Retrieval}} \\
\cline{3-12} 
   & & R@1 & R@5 & R@10 & MedR & Rsum & R@1 & R@5 & R@10 & MedR & Rsum \\
  \hline
  MVS & Base & 0.0 & 0.2 & 0.4 & 1319 & 0.6 & 0.0 & 0.2 & 0.4 & 1314 & 0.6 \\
  MVS &HowTo100M\cite{howto100m} & 24.9 & 30.1 & 32.5 & 147 & 87.5 & 21.8 & 25.7 & 29.5 & 128 & 77.0 \\
  MVS &AVLnet\cite{avlnet} & 18.5 & 31.6 & 38.3 & 44 &88.4& 15.3 & 31.1 & 38.1 & 45 & 84.5 \\
  MVS &MoCo\cite{he2020momentum} & 35.3 & 41.1 & 42.9 & 45 &119.3& 37.0 & 40.8 & 41.1 & 40 & 118.9 \\
  MVS &CLIP\cite{clip} & 35.5 & 42.3 & 43.7 & 77 &121.5& 33.5 & 41.5 & 43.4 & 68 & 118.4 \\
  MVS &Hit\cite{hit} & \underline{40.0} & \underline{42.8} & \underline{44.7} & \underline{35} & \underline{127.5} & \underline{42.7} & \underline{43.9} & \underline{44.9} & \underline{32} & \underline{131.5} \\
  \cline{1-12}
  MVS & MQMC & \textbf{44.7} & \textbf{48.7} & \textbf{50.2} & \textbf{10} & \textbf{143.6} & \textbf{44.9} & \textbf{48.7} & \textbf{50.5} & \textbf{9} & \textbf{144.1} \\
  \hline
  \hline
  MVS & $Improv\%$ & 11.75\% & 13.79\% & 12.30\% & 71.43\% & 12.63\% & 5.15\% & 10.93\% & 12.47\% & 71.88\% & 9.58\% \\
  \cline{1-12}
  \midrule
  
  MVS-large & Base & 0.0 & 0.0 & 0.0 & 12643 & 0.0 & 0.0 & 0.0 & 0.0 & 12734 & 0.0 \\
  MVS-large &HowTo100M\cite{howto100m} & 7.7 & 20.1 & 27.1 & 83 & 54.9 & 7.5 & 19.8 & 26.7 & 107 & 54.0 \\
  MVS-large &AVLnet\cite{avlnet} & 16.8 & 38.0 & 46.6 & 14 & 101.4 & 17.0 & 37.6 & 46.2 & 14 & 100.8 \\
  MVS-large &CLIP\cite{clip} & 20.8 & 43.6 & 51.0 & 10 & 115.4 & 18.4 & 38.9 & 47.0 & 14 & 104.3 \\  
  MVS-large &MoCo\cite{he2020momentum} & 21.2 & 44.0 & \underline{51.3} & \underline{9} &116.5& 19.1 & 39.5 & 47.7 & 13 & 106.3 \\
  MVS-large &Hit\cite{hit} & \underline{24.5} & \underline{44.1} & 50.4 & 10 & \underline{119.0} & \underline{20.9} & \underline{41.2} & \underline{49.5} & \underline{11} & \underline{111.6} \\
  \cline{1-12}
  MVS-large & MQMC & \textbf{27.3} & \textbf{47.7} & \textbf{54.2} & \textbf{7} & \textbf{129.2}  & \textbf{25.1} & \textbf{46.6} & \textbf{53.9} &  \textbf{7} & \textbf{125.5} \\
  \hline
  \hline
  MVS-large & $Improv\%$ & 11.43\% & 8.16\% & 5.65\% & 22.22\% & 8.57\% & 20.10\% & 13.11\% & 8.89\% & 36.36\% & 12.46\% \\
  \bottomrule
\end{tabular}
\end{table*}

\section{Experiments}
To evaluate our proposed model, we conduct extensive experiments on two constructed datasets. In this section, we elaborate on the datasets, metrics, and baselines for evaluation followed by the description of implementation details. 
\subsection{Datasets}
To test our proposed model, we collect two large-scale datasets, MVS and MVS-large, from the micro-video sharing platform. They consist of some microvideo-product pairs, which are embedded by their publishers manually. 
In particular, we capture the micro-video content information, including the micro-videos and their captions. As for the products, we crawl both their pictures associated with textual descriptions. 
Beyond the content information, we manually construct a hierarchical category ontology according to the categories of the products. 

Finally, we obtain the MVS dataset with $13,165$ microvideo-product pairs and MVS-large with $126,206$ pairs. Moreover, we achieve a three-layer category ontology, ranging from clothes to furniture, consisting of $6$ coarse-grained categories, $30$ middle-grained categories, and $316$ fine-grained categories\footnote{Due to insufficient instances of fine-grained categories, we conduct our experiments with other two-level categories.}. 
As shown in Figure~\ref{fig2}, we present the statistics of partial data of the hierarchical category ontology.

\subsection{Evaluation Metrics}

For each dataset, we used the ratio 3:1:1 to randomly split the microvideo-product pairs to constitute the train set, validation set, and test set. The train set, the validation set, and the test set are used to optimize parameters, tune the hyper-parameters and evaluate the performance in the experiments, respectively.

We test the performance with several metrics widely used in information retrieval, including Recall at K (\textit{i.e.,} R@K and K=1, 5, 10), Median Rank (MedR), and Rsum. Specifically, R@K is the percentage of test queries that the relevant item found among the top-K retrieved results. The MedR measures the median rank of correct items in the retrieved ranking list, where a lower score indicates a better model. We also take the sum of all R@K as Rsum to reflect the overall retrieval performance. 

\begin{figure*}
\centering
    \setlength{\abovecaptionskip}{0.2cm}
    \setlength{\belowcaptionskip}{0.cm}
\vspace{-0.5cm} 
	\subfigtopskip=2pt 
	\subfigbottomskip=0pt 
	\subfigcapskip=-2pt 
	\subfigure[MVS (Microvideo-Product)]{
		\label{level.sub.1}
		\includegraphics[width=0.49\linewidth]{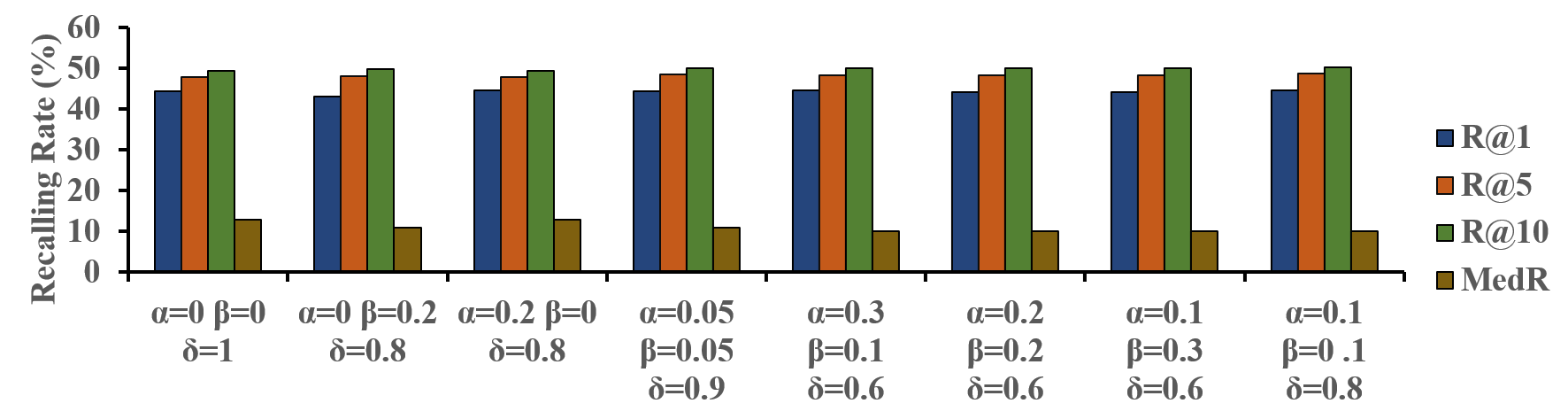}}
	\subfigure[MVS (Product-Microvideo)]{
		\label{level.sub.2}
		\includegraphics[width=0.49\linewidth]{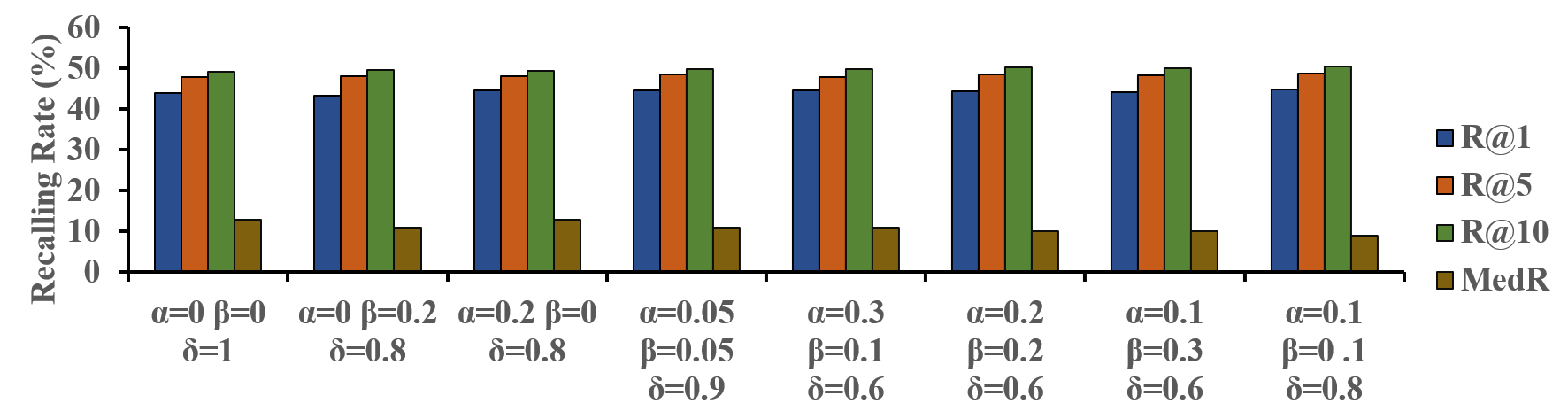}}
	\subfigure[MVS-large (Microvideo-Product)]{
		\label{level.sub.1}
		\includegraphics[width=0.49\linewidth]{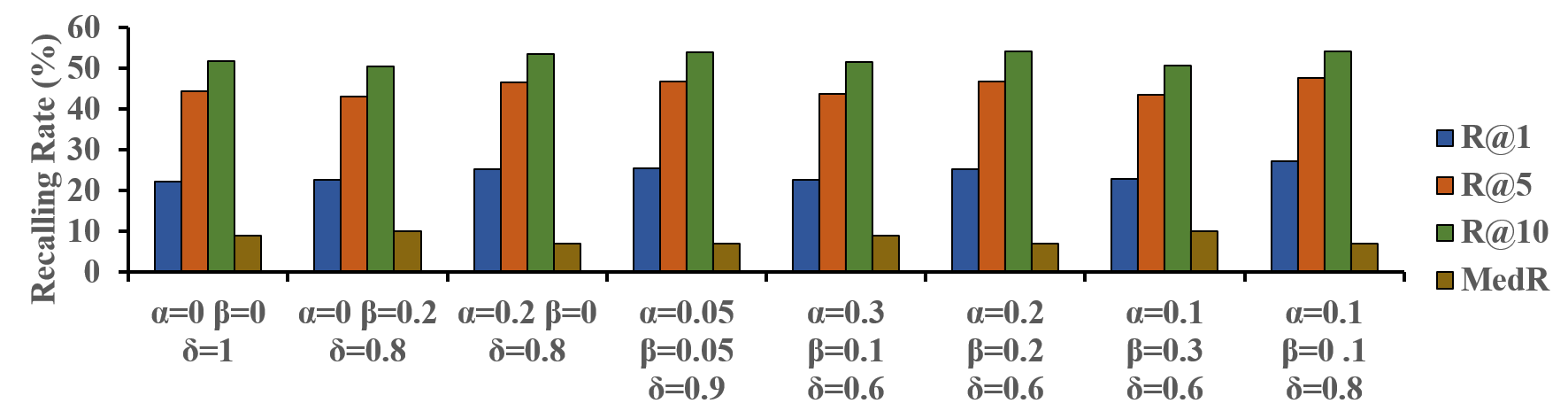}}
	\subfigure[MVS-large (Product-Microvideo)]{
		\label{level.sub.2}
		\includegraphics[width=0.49\linewidth]{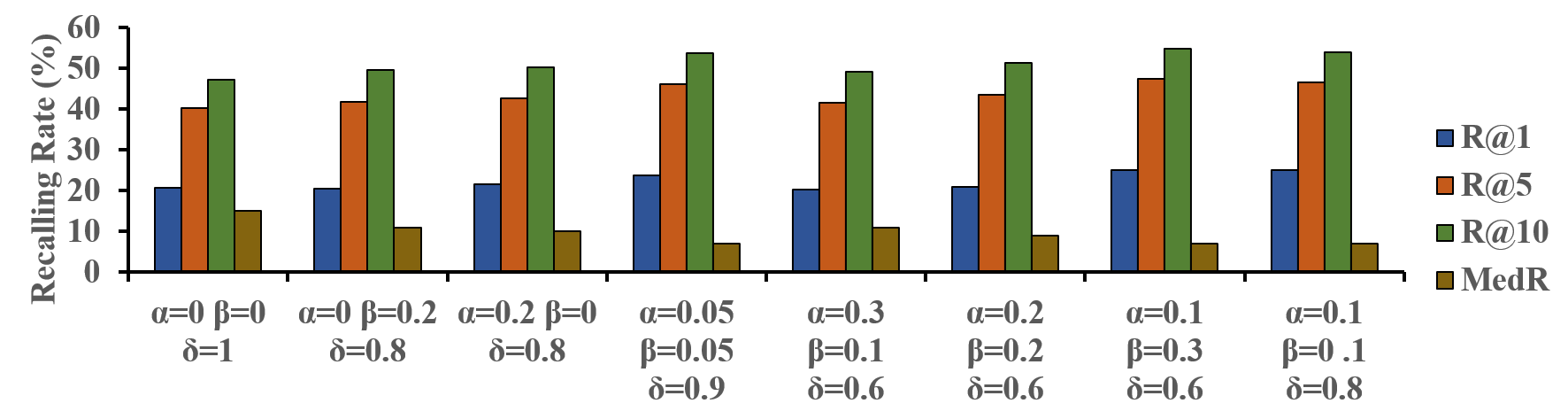}}
\caption{Ablation study on different contrasts.}
\label{Contrast Utilization}
\end{figure*}
\subsection{baselines}
We compare our proposed model with state-of-the-art 
models, including
\begin{itemize}[leftmargin=*]
\item \textbf{Base} The benchmark method concatenates visual and textual features of an instance to measure similarity.
\item \textbf{HowTo100M}\cite{howto100m} This method learns a joint text-video embedding from the paired videos and captions and uses the max-margin ranking loss with a negative sampling strategy.
\item \textbf{AVLnet}\cite{avlnet} This method introduces a self-supervised network that learns a shared audio-visual embedding space from raw audio, video, and text for audio-video retrieval.
\item \textbf{MoCo}\cite{he2020momentum} This method introduces momentum contrast for unsupervised visual embedding. This method keeps the dictionary keys as consistent as possible despite its evolution under the hypothesis that good representation benefits from a large dictionary containing a rich set of negative samples.
\item \textbf{CLIP}\cite{clip} This method learns a multi-modal embedding space of image and text by contrastive pre-training on large-scale datasets. This method maximizes the cosine similarity of the image and text embeddings of the N real pairs in the batch and minimizes the cosine similarity of the other $N^2 - N$ incorrect pairings. 
\item \textbf{Hit}\cite{hit} This method learns hierarchical embeddings with a hierarchical transformer for video-text retrieval, which performs hierarchical cross-modal contrastive matching at both feature-level and semantic-level.
\end{itemize}

\subsection{Implementation Details}
We adopt pre-trained feature extractors in different modalities. In particular, we extract 2,048-dimensional visual features with Resnet152 model~\cite{resnet152} pre-trained on ImageNet and 768-dimensional textual features with BERT-base-uncased model~\cite{bert} pre-trained on the wiki. The visual features of multiple frames are pooled as the feature of one micro-video. For AVLnet~\cite{avlnet}, we also extracted the 128-dimensional audio features from VGGish model~\cite{vgg} pre-trained on YT8M~\cite{youtube8m}. 

We set the hidden size of visual and textual projectors to 1,024. The dimensions of micro-video and product encoding are both set to 512. The leaky ReLU\cite{leakyrelu} is used as the activation function and BatchNorm is appended to hidden layers. The initial learning rate is set to $1e-4$ and the network is optimized by Adam~\cite{adam} optimizer. The weight decay is set to $1e-3$ and cosine decay is used for scheduling the learning rate. The momentum of updating the key encoder is set to 0.999 and $\tau$ is set to 0.07. The length of multi-queue, \textit{i.e.,} $T$, should vary with the batch size. we set the batch size as $64$ and $256$ on MVS and MVS-large datasets, respectively. And, $T$ of MVS and MVS-large are set as $192$ and $2,048$, respectively. For the baselines, we do the same options and follow the designs in their articles to achieve the best performance.

\subsection{Overall Performance Comparison}
To demonstrate the effectiveness of our proposed model, we start by doing a comparison between our proposed model and the baselines on MSV and MSV-large datasets, respectively. Specifically, we list their results \textit{w.r.t} recall and MedR in Table \ref{SOTA}, where $Improv.\%$ represents the relative improvement of the best-performing method (bolded) over the strongest baselines (underlined). 
\begin{itemize}[leftmargin=*]
    \item Without any doubt, our proposed model outperforms all comparison methods by a clear margin. In particular, our proposed model improves over the strongest baselines \textit{w.r.t.} R@0 by 3.79\% and 6.80\% in MVS and MVS-large datasets, respectively. It demonstrates the effectiveness of our proposed model. 
    \item Comparing with the other contrastive-based model (\textit{i.e.,} MoCo, Clip), we find that our proposed model achieves better performance. We attribute such an improvement to our multi-queue contrastive training strategy. 
    \item Beyond the visual and textual information, MQMC considers the acoustic features in the microvideo-product retrieval yet unexpectedly performances are poor in most cases. The reason might be that the content of the micro-video contains some noise information that negatively affects the performance. It verifies our arguments and the reasonability of our proposed model again. 
\end{itemize}



\begin{table}
\vspace{-0.3cm}
    \setlength{\abovecaptionskip}{0.2cm}
    \setlength{\belowcaptionskip}{0.cm}
\caption{Ablation study on different modalities.}
\label{modality ablation} 
\resizebox{\linewidth}{!}{
\begin{tabular}{c|c|ccccc}
\toprule
\multirow{2}{*}{\textbf{Datasets}}
& \multirow{2}{*}{\textbf{Modalities}}
& \multicolumn{5}{c}{\textbf{Microvideo-Product Retrieval}} \\
\cline{3-7} 
& & R@1 & R@5 & R@10 & MedR & Rsum \\
\hline
  MVS & Visual only & 36.9 & 43.5 & 44.3 & 20 &124.7\\
  MVS & Text only & 26.1 & 37.0 & 41.8 & 31 & 104.9\\
  MVS & \textbf{All}  & \textbf{44.7} & \textbf{48.7} & \textbf{50.2} & \textbf{10} & \textbf{143.6} \\
  \cline{1-7}
  \midrule
  MVS-large & Visual only & 17.8 & 36.7 & 44.5 & 18 & 98.9 \\
  MVS-large & Text only & 10.9 &	28.6 &	36.7 &	35 & 76.3\\
  MVS-large & \textbf{All} & \textbf{27.3} & \textbf{47.7} & \textbf{54.2} & \textbf{7} & \textbf{129.2}\\
  
  \bottomrule
\end{tabular}}
\end{table}

\subsection{Ablation Study}
To evaluate the designs in our proposed model, we conduct ablation studies on the two datasets. We test the effectiveness of uni-modal features and multi-modal instance representation learning. In addition, we evaluate the hyper-parameters $\alpha$, $\beta$, and $\delta$ designed for the balance between the representation learning parts. Next, we dive into the multi-queue contrastive training strategy to further test its effectiveness and robustness. 


\begin{table}
\vspace{-0.3cm}
    \setlength{\abovecaptionskip}{0.2cm}
    \setlength{\belowcaptionskip}{0.cm}
\caption{The impacts of Multi-queue for retrieval performance. Sing. and Multi. denote the single and multiple queues, respectively. w/o S. represents the queues without important scores. }
\label{Multi-queue Utilization} 
\resizebox{\linewidth}{!}{
\begin{tabular}{c|c|ccccc}
\toprule
\multirow{2}{*}{\textbf{Datasets}}
& \multirow{2}{*}{\textbf{Queue}}
& \multicolumn{5}{c}{\textbf{Microvideo-Product Retrieval}} \\
\cline{3-7} 
  &  & R@1 & R@5 & R@10 & MedR & Rsum \\
  \hline
  MVS & Sing. \& w/o S. & 42.5 & 47.1 & 48.3 & 17 & 137.9 \\
  MVS & Multi. \& w/o S. &  43.8 & 47.4 & 48.8 & 18 & 140.0 \\
  MVS & Multi. & \textbf{44.7} & \textbf{48.7} & \textbf{50.2} & \textbf{10} & \textbf{143.6} \\
  \cline{1-7}
  \midrule
  MVS-large & Sing.\& w/o S. & 18.5 & 38.5 & 46.3 & 15 & 103.3 \\
  MVS-large & Multi.\& w/o S. & 21.4 & 41.7 & 49.0 & 12 & 112.1 \\
  MVS-large & Multi. & \textbf{27.3} & \textbf{47.7} & \textbf{54.2} & \textbf{7} & \textbf{129.2} \\
  \bottomrule
\end{tabular}}
\end{table}

\subsubsection{\textbf{Uni-modal and multi-modal Representation Learning}}
In order to test the uni-modal and multi-modal representation learning, we first explore the effects of different modalities and compare the results over the two datasets, as listed in Table~\ref{modality ablation}. From the results, we observe that:
\begin{itemize}[leftmargin=*]
    \item As expected, the performance with multi-modal representation learning significantly outperforms that with uni-modal ones, including visual and textual modalities, on MVS and MVS-large datasets. It demonstrates the effectiveness of the multi-modal instance representation learning, including the multi-modal fusion method and cross-instance contrastive loss.
    \item Jointly analyzing the results of the baselines model shown in Table~\ref{SOTA}, we find that the performance with uni-modal representation learning is even better than that of the state-of-the-art baselines in some cases. We attribute this phenomenon to our designed uni-modal representation learning, which distills the informative signal from the noise information caused by the complex and chaotic background. 
\end{itemize}

Next, we perform the experiments by varying the value of hyper-parameters $\alpha$, $\beta$, and $\delta$ in the range of 0 to 1. Observing the results illustrated in Figure~\ref{Contrast Utilization}, we have the following findings the cross-instance contrast plays a vital role in the microvideo-product retrieval task. This might be that the supervision signal from the cross-instance similarity makes more contribution to optimize the uni- and multi-modal representation learning. Nevertheless, the cross- and intra-modal contrastive losses also cannot be ignored, which verifies our uni-modal feature representation learning again. 
Overall, from the results, we find that our proposed model gains the best performance when we set $\alpha=0.1$, $\beta=0.1$, and $\delta=0.8$, respectively.

\begin{figure}
\centering
    \setlength{\abovecaptionskip}{0.1cm}
    \setlength{\belowcaptionskip}{0.cm}
\vspace{-0.5cm} 
	\subfigtopskip=2pt 
	\subfigbottomskip=2pt 
	\subfigcapskip=-3pt 
	\subfigure[MVS]{
		\label{level.sub.1}
		\includegraphics[width=0.45\linewidth]{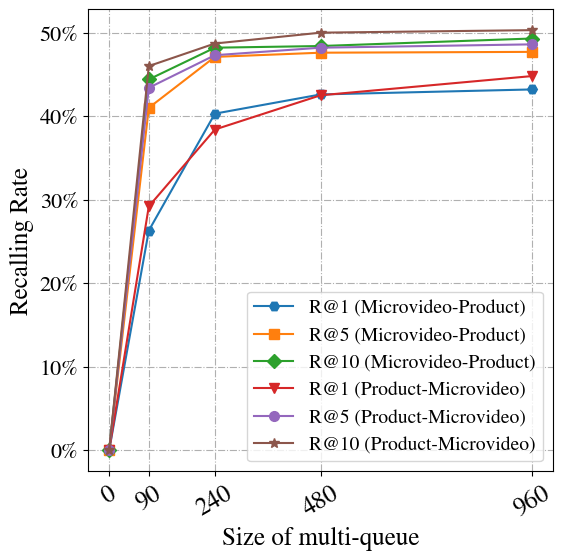}}
	\subfigure[MVS-large]{
		\label{level.sub.2}
		\includegraphics[width=0.45\linewidth]{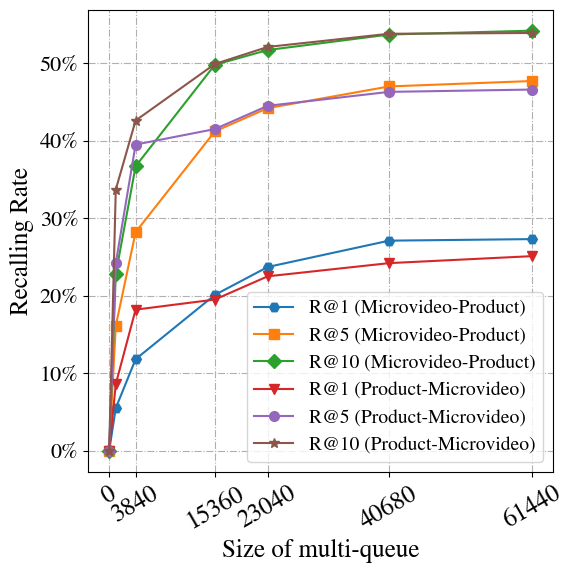}}
\caption{Ablation study on two datasets to investigate the contributions of different sizes of Multi-queue.}
\label{mk}
\end{figure}
\subsubsection{\textbf{Multi-queue Contrastive Training Strategy}}
To evaluate the effectiveness of the multi-queue, we conduct the experiments of our proposed model without the multi-queue architecture and compare its performance with that of our proposed model equipped with the multi-queue. Moreover, we discard the difference of multiple queues by setting the weight (\textit{i.e.,} $e_{i,j}$) of each anchor-negative pair to 1. We list the results in Table~\ref{Multi-queue Utilization} and find that: 
the results \textit{w.r.t.} R@1, R@5, R@10, and MedR are increased when we utilize the multi-queue architecture. It justifies our designed multi-queue contrastive loss. 
Furthermore, we compare the performance with the importance scores and without the scores. The performance \textit{w.r.t.} R@1 is improved from 43.8\% to 44.7\% on MVS and from 21.4 \% to 27.3\% on MVS-large. We suggest that the proposed model is benefited from the utilization of the scores to distinguish the importance of different negatives.  

Further, we explore the influences of the multi-queue size in microvideo-product retrieval. For this goal, we conduct the experiments on the two datasets by varying the sizes from 90 to 960 and from 960 to 61440, respectively. Observing the experimental results shown in Figure~\ref{mk}, we find that the introduction of large-scale negatives for similarity learning indeed achieves considerable performance improvements. We attribute it to broader negative interactions for obtaining more precise and discriminative representations. In fact, due to the unbalanced distribution of categories and the existence of the long-tail problem, the size of the multi-queue is limited by a few categories with a small amount of data. When the actual training samples cannot fill the queue, the negatives in the queue are not single and independent, so the repeated positive cases in the queue are mistakenly divided into negative cases. Moreover, by reason of the category sensitivity of multi-queue, the above errors will be magnified, affecting the learning effect of the retrieval modal.


\subsection{Case Study}
\begin{figure}
\vspace{-0.5cm}
\centering 
    \setlength{\abovecaptionskip}{0.1cm}
    \setlength{\belowcaptionskip}{0.cm}
\includegraphics[width=0.48\textwidth]{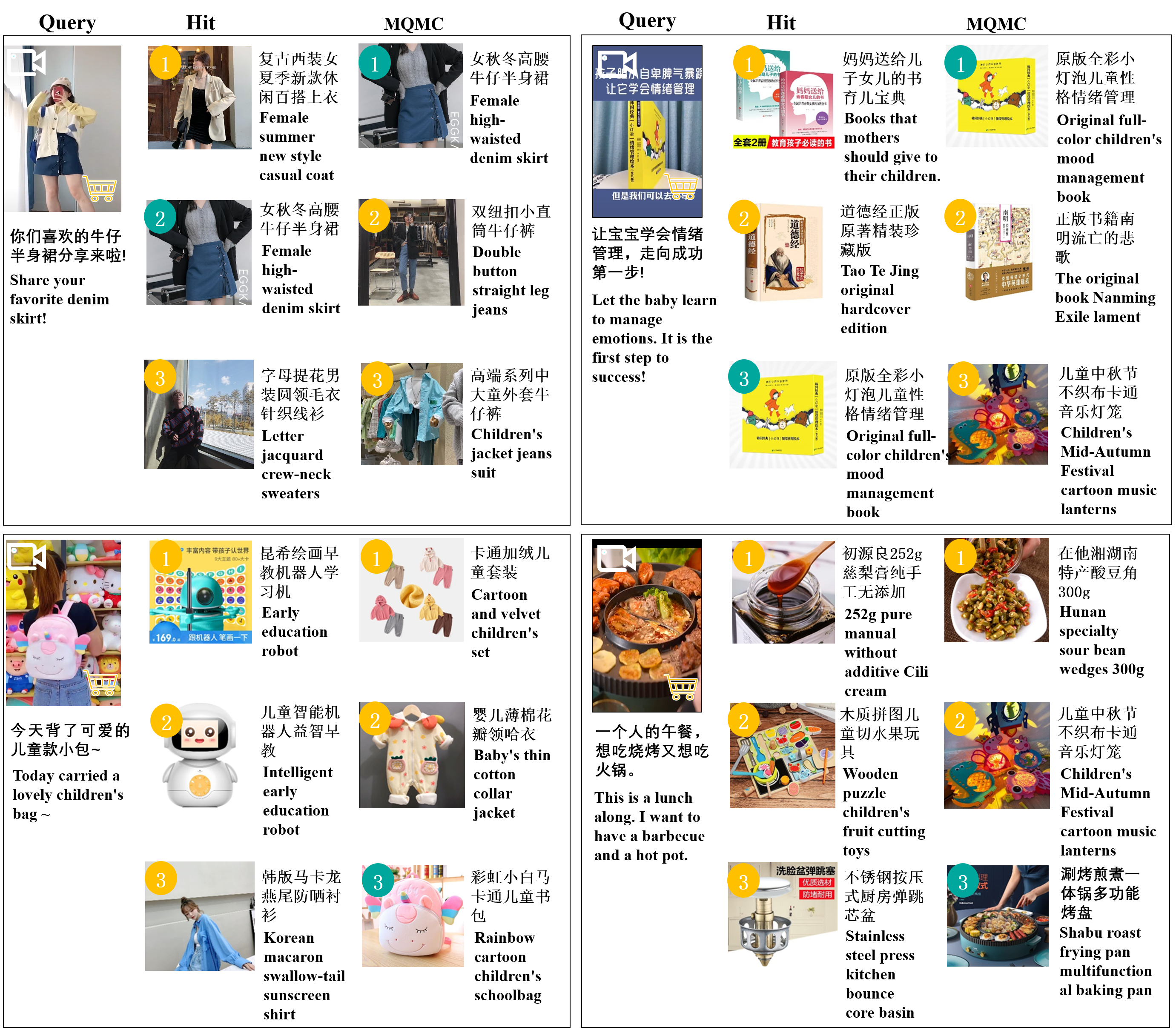}
\caption{Four examples with top-3 microvideo-product retrieval results of MQMC and Hit.}
\label{case}
\end{figure}

To visualize our proposed model, we randomly sample four micro-videos from two datasets and conduct the strongest baseline (Hit) and our proposed models on them. In particular, we collect the Top-3 results of two models~\ref{case} and separately mark the correct predictions with green circles and others with yellow circles at the top left corners, as shown in Figure~\ref{case}.

According to the results, in the first two cases, our proposed model not only matches the visually similar items in micro-videos and pictures of products but also explores the relationships between textual and visual modalities sufficiently to obtain identical ones more accurately. In the upper left case, with the assistance of textual modality, confusing visual modality can focus more attention on the skirt that the micro-video wants to display, rather than the coat that occupies more space in the frame. In the upper right case, our proposed model can effectively remove the interference of large areas of the blue background and white characters in videos, and perform accurate retrieval for book categories with fewer data. However, in the last two cases, due to the obvious visual differences between some micro-videos and products, our proposed model can not achieve the best match. In the lower-left case, the video background is very cluttered, with multiple indistinguishable objects that are very similar to each other. In the lower right case, due to a large amount of variant visual information between the frame and picture, as well as the incomplete outline of the item limited by the shooting angle of the micro-video, the retrieval faces great difficulty.

\section{Conclusion}
This paper is the first to formulate the microvideo-product retrieval task between multi-modal and multi-modal instances.
We propose the Multi-Queue Momentum Contrastive learning network (MQMC) for bidirectional microvideo-product multimodal-to-multimodal retrieval, which consists of the uni-model feature representation learning and multi-modal instance representation learning, integrating cross-modal contrast, intra-modal contrast and cross-instance contrast organically. 
In addition, we present a discriminative negative selection strategy with a multi-queue to distinguish the importance of different negatives with their categories.  
Two large-scale multi-modal datasets are built for microvideo-product retrieval, and we construct a category ontology and manually annotate the multi-layer ontologies of all products of the datasets. Extensive experiments prove the validity of the proposed model.
In the future, MQMC can be extended as a general method for multimodal-to-multimodal retrieval. And more technological innovations can be researched such as multi-modal integrated approaches, hard negative selection strategy, etc.

\clearpage

\bibliographystyle{ACM-Reference-Format}
\bibliography{sample-base}

\end{document}